\documentclass[10pt,twocolumn]{article}

\usepackage[letterpaper,top=0.68in,bottom=0.72in,left=0.67in,right=0.67in,columnsep=0.24in]{geometry}
\usepackage[T1]{fontenc}
\usepackage{textcomp}
\usepackage{newtxtext,newtxmath}
\usepackage{graphicx}
\usepackage{amsmath}
\usepackage{booktabs}
\usepackage{balance}
\usepackage{tabularx}
\usepackage{array}
\usepackage{multirow}
\usepackage{enumitem}
\usepackage{xcolor}
\usepackage{microtype}
\usepackage{xspace}
\usepackage[font=small,labelfont=bf,labelsep=period]{caption}
\usepackage{subcaption}
\usepackage{placeins}
\usepackage{dblfloatfix}
\usepackage{balance}
\usepackage{listings}
\usepackage{float}
\usepackage[most]{tcolorbox}
\usepackage{titlesec}
\usepackage[
  backend=biber,
  style=numeric-comp,
  sorting=none,
  sortcites=true,
  defernumbers=true,
  maxbibnames=99,
  giveninits=true,
  natbib=true,
  doi=true,
  url=false,
  isbn=false
]{biblatex}
\usepackage[hidelinks]{hyperref}

\hypersetup{
  pdftitle={Rethinking Foundation Model Collaboration: Enhancing Specialized Models via Proxy Task Reasoning},
  pdfauthor={Hongyi Lin, Fei Ma, Ziyu Lin, Song Zhang, Yang Liu, Xiaobo Qu, and Jinhua Zhao}
}


\setlist[itemize]{leftmargin=1.35em,itemsep=1pt,topsep=2pt}
\setlist[enumerate]{leftmargin=1.5em,itemsep=1pt,topsep=2pt}
\newcolumntype{Y}{>{\raggedright\arraybackslash}X}
\newcolumntype{C}{>{\centering\arraybackslash}X}
\newcommand{\framework}{FAT\xspace}
\newcommand{\method}{ProxySelect\xspace}

\newcommand{\best}[1]{\textbf{#1}}

\lstdefinestyle{promptstyle}{
  basicstyle=\rmfamily\fontsize{6.45}{6.8}\selectfont,
  breaklines=true,
  breakatwhitespace=true,
  columns=fullflexible,
  keepspaces=true,
  showstringspaces=false,
  upquote=true,
  aboveskip=0pt,
  belowskip=0pt
}

\newtcblisting{promptbox}[1]{
  enhanced,
  breakable,
  listing only,
  listing engine=listings,
  listing options={style=promptstyle},
  colback=black!2,
  colframe=black!35,
  coltitle=black,
  colbacktitle=black!7,
  boxrule=0.45pt,
  arc=1.5pt,
  title={#1},
  fonttitle=\bfseries\scriptsize,
  left=1.0mm,
  right=1.0mm,
  top=0.25mm,
  bottom=0.25mm,
  before skip=0.12em,
  after skip=0.12em
}

\title{\vspace{-0.8em}\textbf{Rethinking Foundation Model Collaboration:\\[-0.08em] Enhancing Specialized Models through Proxy Task Reasoning}}
\author{
\ Hongyi Lin$^{1,2}$, Yang Liu$^{1}$, Jinhua Zhao$^{2}$, and Xiaobo Qu$^{1}$\\[0.35em]
\small $^{1}$School of Vehicle and Mobility, Tsinghua University, Beijing, China\\
\small $^{2}$Department of Urban Studies and Planning, Massachusetts Institute of Technology, Cambridge, MA, USA
}
\date{}

\begin{document}
\flushbottom
\maketitle
\vspace{-1.2em}
\begin{refsection}

\begin{abstract}
Foundation models are increasingly integrated into embodied intelligence systems, but directly assigning them structured prediction tasks requires precise geometric and numerical estimation, where specialized models often remain stronger. This capability mismatch raises a key question: should foundation models replace task-specific predictors, or should they collaborate through tasks better aligned with their strengths? We propose \framework, a foundation-model-augmented task-specific reasoning framework that treats collaboration as task decomposition rather than model replacement. \framework decomposes structured prediction into specialist prediction, information-space reconstruction, and foundation-model proxy reasoning. The specialist generates geometrically and physically valid hypotheses in the native output space, while the foundation model performs a bounded proxy task, such as selection or verification, over reconstructed multimodal candidates. We instantiate this principle as \method with a vision--language model. Across 2D object detection, 3D object detection, trajectory prediction, and semantic segmentation, \method consistently improves specialized baselines and substantially outperforms direct foundation-model regression at lower computational cost. These results suggest a general collaboration principle: specialized models preserve task-specific structure, while foundation models refine their hypotheses through contextual proxy reasoning.
\end{abstract}

\section{Introduction}

Foundation models are becoming general interfaces for perception, reasoning, and decision support in embodied intelligence systems. In transportation, robotics, and autonomous driving, large multimodal models have been used for knowledge organization, scene understanding, planning assistance, and language-conditioned decision making \citep{qu2023envisioning,kuang2024harnessing,hu2025multimodal,zhou2024survey,tian2025drivevlm,monwilliams2025}. Their broad visual and linguistic priors make them effective at interpreting context, recognizing unusual situations, and connecting observations with task instructions. However, many embodied intelligence tasks are not purely semantic. A 3D bounding box, a dense segmentation mask, or a future trajectory is a structured output with strict geometric, physical, and numerical requirements. Small depth or yaw errors can change obstacle localization; discontinuous waypoints can imply infeasible motion; and boundary leakage can affect free-space estimation.

This creates a fundamental question for foundation-model collaboration: should a foundation model directly solve the original specialized task, or should it be assigned a role better aligned with its strengths? Existing integration strategies often treat foundation models as direct predictors or end-to-end replacements for task-specific modules. Although this design benefits from generalization, it also forces the foundation model to perform precise metric estimation, structured decoding, and domain-specific calibration, where carefully trained specialized models often remain stronger. Consequently, the complementary strengths of the two models are not fully utilized: specialized models are strong at precise numerical estimation, whereas foundation models are naturally stronger at semantic understanding, contextual reasoning, and comparative judgment.

We argue that effective collaboration should arise not from replacing specialized models, but from assigning each model the task that best matches its capability. Direct structured prediction requires searching a high-dimensional continuous output space, whereas many decision processes surrounding structured prediction can be reformulated as bounded reasoning problems over a finite set of hypotheses. Tasks such as comparison, verification, ranking, consistency checking, and ambiguity resolution largely rely on semantic and contextual reasoning instead of precise metric regression, making them substantially better aligned with the strengths of foundation models.

Our intuition is inspired by human annotation workflows. When correcting machine-generated results, annotators rarely redo the original prediction from scratch. Instead, they compare candidate boxes, verify segmentation boundaries, select plausible trajectories, or refine uncertain regions. In this process, a difficult structured prediction problem is converted into a simpler proxy reasoning task. This decomposition is one reason human annotation can efficiently produce high-quality labels: quality is achieved through comparison, verification, and refinement over structured hypotheses, rather than repeated prediction from scratch. This observation suggests that foundation models may contribute more effectively by reasoning over structured hypotheses than by directly generating them.

Based on this hypothesis, we propose a Foundation-Model-Augmented Task-Specific Reasoning framework, termed \framework. The framework decomposes the original prediction process into three stages: specialized-model prediction, information-space reconstruction, and foundation-model proxy reasoning. First, a task-specific model generates geometrically or physically grounded hypotheses in its native output space. Second, these hypotheses are reconstructed into a foundation-model-readable information space through overlays, projections, structured descriptions, or region identifiers. Third, the foundation model solves a bounded proxy task, such as selection, verification, comparison, or consistency checking. The final output is mapped back to the specialized model's structured output space, rather than being freely generated by the foundation model.

We instantiate this framework as \method, a vision--language proxy-task learner for structured prediction. In this instantiation, the specialized model generates candidate boxes, trajectories, or mask regions; these candidates are rendered into 2D image-based multimodal representations; and the VLM selects or verifies the most plausible candidate. The comparison with direct VLM regression is therefore a comparison between two collaboration paradigms: using the foundation model as the primary structured regressor, or using it as a proxy reasoner that improves specialist-generated hypotheses.

Our contributions are threefold:
\begin{itemize}
  \item We propose \framework, which reframes foundation-model collaboration as proxy-task reasoning rather than model replacement.
  \item We instantiate \framework as \method, mapping boxes, trajectories, and masks into a unified multimodal interface for selection and verification.
  \item We validate the framework across four embodied intelligence tasks, showing consistent gains over specialists and better efficiency than direct foundation-model regression.
\end{itemize}

\section{Related Work}

\paragraph{Generalist--specialist collaboration.}
Recent autonomous-driving and robotics systems increasingly combine multimodal foundation models with task-specific modules. DriveVLM integrates VLM-based scene description, scene analysis, and hierarchical planning for autonomous driving \citep{tian2025drivevlm}; GRAPE uses preference alignment and VLM-proposed constraints to improve robot policies \citep{zhang2024grape}; and VLM-AD uses a foundation model as an offline supervisor for end-to-end driving \citep{xu2025vlmad}. Complementary perception studies retain specialist metric modules for BEV fusion, sensor calibration, and novel-object detection \citep{lin2025aefusion,lin2025calibration,zhuang2025fewshot}, while scene-description supervision has been explored for end-to-end autonomous driving \citep{liu2025desead}. These studies demonstrate the value of collaboration between general-purpose models and specialized modules. \framework differs by making this collaboration explicit as proxy-task reasoning: the specialized model constructs task-valid hypotheses, the representation module reconstructs them into a foundation-model-readable information space, and the foundation model performs bounded reasoning over the reconstructed alternatives.

\paragraph{Structured prediction with multimodal models.}
Multimodal and grounded vision-language models can emit or support structured visual outputs such as boxes, points, masks, and grounded object regions \citep{yang2023gpt4v,bai2025qwen,liu2024groundingdino}. Sequence-based detection and reasoning segmentation further demonstrate the flexibility of generative or language-conditioned structured prediction \citep{chen2022pix2seq,lai2024lisa}. Nevertheless, directly decoding numerical structures remains challenging when outputs must satisfy tight coordinate conventions, physical continuity, or dense-boundary constraints. Our focus is therefore not whether a foundation model can be fine-tuned to generate structured outputs, but whether direct generation is the most reliable allocation of responsibility in embodied intelligence systems.

\paragraph{Candidate generation and comparative reasoning.}
Candidate generation and reranking are established tools in structured prediction. Motion forecasting methods such as MultiPath and CoverNet represent future motion with finite trajectory sets \citep{chai2019multipath,philion2020covernet}, while HPNet produces multimodal futures through historical prediction attention \citep{tang2024hpnet}. Generative transportation research has progressed from GAN-based augmentation \citep{lin2023ganits} to large-model-guided and risk-controllable driving scenario synthesis \citep{zhang2026dynamic,lin2026riskmv}; these methods enrich the hypothesis space, whereas our work focuses on bounded arbitration over specialist-generated hypotheses. Preference learning and comparative judgment similarly exploit relative evaluation rather than requiring a single unconstrained output \citep{ouyang2022,verhavert2019}. \framework extends this propose--judge pattern to embodied structured prediction by conditioning a foundation model on the original scene and explicit visualizations of specialist hypotheses, with \method serving as a concrete selection- and verification-based instantiation.

\section{Foundation-Model-Augmented Task-Specific Reasoning}
\label{sec:method}

\subsection{Framework overview}

\framework formulates foundation-model collaboration as proxy-task reasoning rather than direct replacement of specialized models. Given an original structured prediction task, the framework decomposes the prediction process into three role-aligned stages.

First, a task-specific model produces an initial prediction or a finite set of hypotheses in its native output space. This stage preserves the geometric, dynamic, or dense spatial constraints learned from domain-specific supervision. Second, an information-space reconstruction module converts these hypotheses into a foundation-model-readable representation, such as image overlays, camera projections, map-conditioned trajectory visualizations, structured candidate descriptions, or numbered mask regions. Third, the foundation model solves a bounded proxy task over the reconstructed information, such as selecting the best candidate, verifying uncertain regions, comparing alternatives, ranking hypotheses, or checking semantic consistency. The proxy decision is then mapped back to the original structured output space.

\begin{figure*}[!t]
  \centering
  \includegraphics[width=\textwidth]{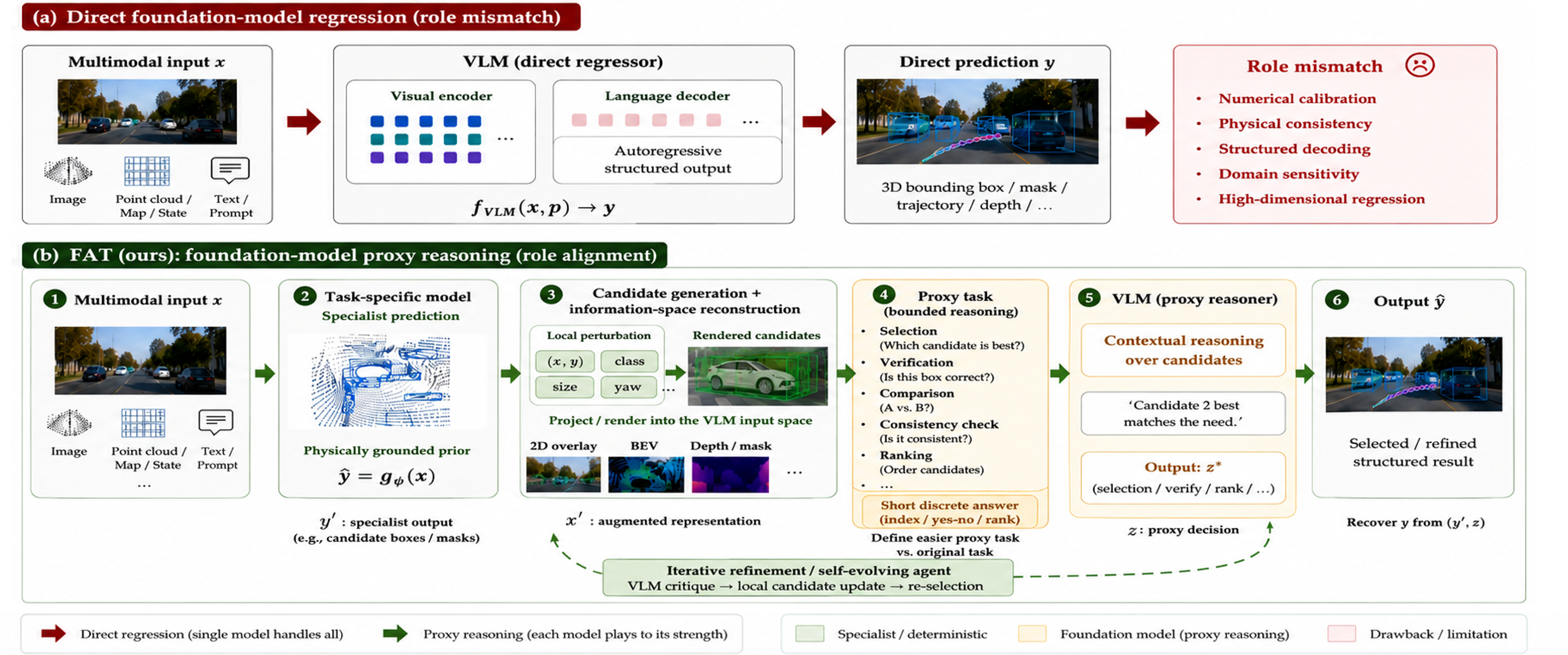}
  \caption{Overview of the \framework framework and its \method instantiation. A task-specific model first generates structurally valid hypotheses. These hypotheses are reconstructed into a foundation-model-readable information space through overlays, projections, structured descriptions, or numbered regions. The foundation model then solves a bounded proxy task, such as selection, verification, comparison, ranking, or consistency checking, and the proxy decision is mapped back to the original structured output space.}
  \label{fig:framework}
\end{figure*}

This decomposition assigns each component a role aligned with its strengths. The specialized model remains responsible for metric structure and physical validity, while the foundation model contributes contextual reasoning and comparative judgment. Fig.~\ref{fig:framework} illustrates this workflow. In the experiments, we instantiate \framework as \method, where the proxy task is implemented mainly as candidate selection or region verification.

\subsection{Probabilistic formulation}

Let $x\in\mathcal{X}$ denote the original multimodal input and $y\in\mathcal{Y}$ denote the desired structured output. A conventional supervised formulation directly models
\begin{equation}
  p_{\theta}(y\mid x),
  \label{eq:direct-prob}
\end{equation}
or learns a deterministic mapping $f_{\theta}:x\mapsto y$. This formulation asks one model to perform semantic reasoning, metric estimation, and structured decoding simultaneously.

\framework instead introduces three intermediate variables: a specialist hypothesis object $y'$, an augmented information representation $x'$, and a proxy decision $a$. The induced prediction distribution can be written as
\begin{equation}
\begin{aligned}
p_{\mathrm{FAT}}(y\mid x,p)
&= \sum_{y'} \int
p_{\eta}(x'\mid x,y')\,p_{\psi}(y'\mid x) \\
&\quad \times
\sum_{a\in\mathcal{A}(y')}
\mathbf{1}\!\left[y=M(y',a)\right] \\
&\quad \times
p_{\phi}(a\mid x,x',y',p)\,d x' .
\end{aligned}
\label{eq:fat-factorization}
\end{equation}
where $p_{\psi}(y'\mid x)$ denotes the task-specific model's prediction or hypothesis-generation process, $p_{\eta}(x'\mid x,y')$ denotes information-space reconstruction, $p_{\phi}(a\mid x,x',y',p)$ denotes foundation-model proxy reasoning under instruction $p$, and $M(y',a)$ maps the proxy decision back to the original structured output space. The action space $\mathcal{A}(y')$ depends on the proxy task: it may contain candidate indices, region subsets, rankings, verification labels, or consistency decisions.

Equation~\eqref{eq:fat-factorization} formalizes the central distinction between direct prediction and proxy-task reasoning. The foundation model is not required to synthesize $y$ from scratch. Instead, it reasons over a reconstructed representation of specialist-generated hypotheses and outputs a bounded proxy decision. In deterministic implementations, Eq.~\eqref{eq:fat-factorization} reduces to a three-stage pipeline:
\begin{equation}
  y'=G_{\psi}(x),\;
  x'=R_{\eta}(x,y'),\;
  \hat{y}=M\!\left(y',H_{\phi}(x,x',y',p)\right).
  \label{eq:deterministic-fat}
\end{equation}

\subsection{\method as a selection and verification instantiation}

In this paper, we instantiate \framework as \method. For most tasks, the specialist produces a finite candidate set
\begin{equation}
  C(x)=\{y_1,\ldots,y_K\},
  \label{eq:candidate-set}
\end{equation}
where each $y_i$ is a valid structured output in the specialist's native output space. The reconstruction module renders or projects the candidates into a multimodal representation
\begin{equation}
  x' = R_{\eta}(x,C(x)).
  \label{eq:reconstruction}
\end{equation}
The proxy action is a candidate identifier $a=i$. The VLM assigns a score to each candidate,
\begin{equation}
  s_i = s_{\phi}(i\mid x,x',C(x),p),
  \label{eq:selection-score}
\end{equation}
and the corresponding categorical probability is
\begin{equation}
  P_{\phi}(i\mid x,x',C(x),p)
  =
  \frac{\exp(s_i)}
  {\sum_{k=1}^{K}\exp(s_k)}.
  \label{eq:selection-prob}
\end{equation}
The selected output is
\begin{equation}
  \hat{i}=\arg\max_{1\le i\le K}P_{\phi}(i\mid x,x',C(x),p),\;
  \hat{y}=M(C(x),\hat{i}) .
  \label{eq:selection-output}
\end{equation}

For supervised proxy-task learning, the target candidate is defined by the task metric,
\begin{equation}
\begin{aligned}
  i^{\mathrm{gt}}
  &=
  \arg\min_i d(y_i,y^{\mathrm{gt}}), \\
  \mathcal{L}_{\mathrm{sel}}
  &=
  -\log P_{\phi}(i^{\mathrm{gt}}\mid x,x',C(x),p),
\end{aligned}
\label{eq:selection-loss}
\end{equation}
where $d(\cdot,\cdot)$ is the task-specific discrepancy, such as box mismatch, displacement error, or mask mismatch. Ground truth is used only to construct training labels and compute evaluation metrics; it is not provided to the foundation model at inference.

Semantic segmentation uses a set-valued proxy decision. The specialist provides a high-confidence semantic prior and a set of atomic regions. The VLM returns a subset of region identifiers, and the mapping back to the output space unions the retained specialist mask with the selected atomic regions:
\begin{equation}
  \hat{m}
  =
  m_{\mathrm{high}}
  \cup
  \bigcup_{j\in S_{\phi}} r_j,
  \label{eq:segmentation-union}
\end{equation}
where $m_{\mathrm{high}}$ is the retained high-confidence specialist mask, $\{r_j\}$ are candidate atomic regions, and $S_{\phi}$ is the VLM-confirmed region subset. Thus, selection and verification are both instances of the same proxy-task formulation.

The direct-regression baseline and \method represent two different foundation-model collaboration paradigms. The direct baseline assigns the foundation model the original structured prediction task, whereas \method assigns it a proxy task over specialist-generated hypotheses. Both use the same Qwen2.5-VL-7B backbone and LoRA adaptation strategy \citep{bai2025qwen,hu2022lora}; the difference lies in the role assigned to the foundation model.

\subsection{Candidate generation and randomization}

The specialist model can be interpreted as defining a posterior distribution over plausible structured outputs. Candidate sets may come from multimodal decoding heads, stochastic predictions, multiple models, or local perturbations. When a specialist provides one prediction $\hat y^{(0)}=g_{\psi}(x)$, we use
\begin{equation}
 C(x)=\{\hat y^{(0)}\}\cup\left\{T_j(\hat y^{(0)};\epsilon_j)\right\}_{j=1}^{K-1},
 \label{eq:candidates}
\end{equation}
where $T_j$ alters task-relevant dimensions such as position, size, orientation, category, speed profile, endpoint, or mask membership. Candidate quality and diversity jointly determine the upper bound available to the selector.

Before rendering each training or evaluation sample, candidates are randomly permuted and renumbered, and labels are remapped after the permutation. No detector proposal, trajectory mode, specialist output, or perturbation source therefore occupies a fixed index. The same randomization is applied independently within each 2D target group and to numbered atomic regions in segmentation.

\subsection{Information-space reconstruction and bounded proxy reasoning}

The information-space reconstruction layer exposes the consequences of each hypothesis to the foundation model. Two-dimensional boxes are overlaid on the image and accompanied by coordinates and classes; 3D boxes are projected into the camera view and described by location, dimensions, yaw, and category; trajectories are drawn against map geometry; and uncertain segmentation regions are numbered. The foundation model performs comparative analysis and returns a candidate identifier or a list of region identifiers. The selected output is mapped back to the corresponding specialist-generated structure. The foundation model never needs to synthesize new coordinates, contours, or motion sequences in the proxy-task pipeline.

A deployment implementation should preserve bounded authority. Invalid indices, malformed outputs, or low-confidence selections can trigger a deterministic fallback to the original specialist result. This fallback is not counted as a claim of safety; it is a practical mechanism that localizes failure and prevents an invalid language-model string from directly entering a downstream control interface.

\subsection{Task instantiations}
\label{sec:task-instantiations}

\paragraph{2D object detection.}
The candidate pool combines RT-DETR and YOLO-family detector outputs \citep{zhao2024rtdetr}. For each RT-DETR anchor, five local perturbations modify center position, width, height, aspect ratio, and class. The VLM sees candidate overlays and a structured list of coordinates and categories, and selection is performed independently within each target group. Every inference-time candidate is a detector output or a perturbation of one; no ground-truth box is used.

\paragraph{3D object detection.}
The specialist is 3D-MOOD \citep{yang2025mood}. Each input contains one specialist box and nine alternatives that perturb 3D position, dimensions, yaw, and category. Wrong-class candidates are included during training. Training candidates are perturbed around annotated boxes, whereas inference candidates are generated only from the specialist prediction.

\paragraph{Trajectory prediction.}
The specialist is HPNet \citep{tang2024hpnet}. Six HPNet modes are augmented with four alternatives containing mild waypoint noise, a gradual direction change, a rapid mid-to-late speed drop, or an incorrect endpoint. The representation includes agent history, lane geometry, a gray inference-time reference trajectory, and colored candidate futures. The gray reference is distinct from the annotated future; the latter is used only for training labels and ADE/FDE evaluation.

\paragraph{Semantic segmentation.}
Mask2Former supplies category-aware predictions and FastSAM supplies atomic geometric partitions \citep{cheng2022mask2former,zhao2023fastsam}. Instances with confidence above $0.500$ form a high-confidence mask retained directly. A FastSAM partition is excluded when at least $0.700$ of it is already covered by that mask or when its uncovered ratio is below $0.100$. Remaining regions are numbered for VLM verification, and the final mask is the union of the high-confidence Mask2Former result and VLM-confirmed atomic regions.

\section{Experiments}

\subsection{Experimental setup}

\paragraph{Datasets.}
We use COCO 2017 \citep{lin2014coco}, KITTI \citep{geiger2013kitti}, Argoverse \citep{chang2019argoverse}, and Cityscapes \citep{cordts2016cityscapes} to cover 2D localization, metric 3D perception, map-conditioned motion forecasting, and dense urban scene understanding, respectively. This selection evaluates the same proxy-task principle across complementary structured outputs and different levels of geometric support.

\paragraph{Models and training.}
The foundation-model component in all four task instantiations uses Qwen2.5-VL-7B with LoRA fine-tuning. Training was conducted on two NVIDIA A100 GPUs. The direct-regression and proxy-task variants use the same VLM backbone and adaptation strategy; they differ only in the role assigned to the foundation model and in the information provided to it. Candidate order is randomized for every sample. Table~\ref{tab:setup} summarizes the task-specific protocols. The main text reports normalized system-level cost for compact comparison, while complete task-specific metrics and absolute GFLOPs per sample are reported in Appendix~\ref{app:full-results}. Absolute GFLOPs are rounded to the nearest integer, and performance metrics are reported to three decimals.

\begin{table*}[!t]
\centering
\caption{Task-specific experimental protocols.}
\label{tab:setup}
\footnotesize
\setlength{\tabcolsep}{4.2pt}
\renewcommand{\arraystretch}{1.12}
\begin{tabularx}{\textwidth}{@{}l Y l Y@{}}
\toprule
Task & Candidate pool and selector output & Fine-tuning data & Evaluation \\
\midrule
2D detection & RT-DETR anchor, five local perturbations, and YOLO-family proposals; one selected box per target group & 118,000 images & COCO 2017; AP and AR \\
3D detection & One 3D-MOOD box and nine perturbations of position, dimensions, yaw, and class; one selected 3D box & 3,000 KITTI images & KITTI; AP and AP$_{50}$ \\
Trajectory prediction & Six HPNet modes and four perturbed alternatives; one selected trajectory index & 15,000 trajectories & Argoverse evaluation set (10,000 trajectories); ADE and FDE \\
Semantic segmentation & Mask2Former high-confidence prior and filtered FastSAM atomic partitions; selected region indices & 14,115 Cityscapes samples & Cityscapes validation set, car category; mIoU, mPrecision, and mRecall \\
\bottomrule
\end{tabularx}
\end{table*}

\subsection{Main results}

Table~\ref{tab:main-results} reports one primary metric per task, following the same task order and metric definitions as the ablations in Table~\ref{tab:source-ablation}. Direct foundation-model regression is consistently less accurate than the corresponding specialist and incurs substantially higher inference cost. In contrast, \method improves every specialist baseline with only a modest cost increase, indicating that assigning the foundation model a bounded proxy task is more effective than using it as a direct structured regressor.

For 2D detection, AP$_{50}$ improves from 0.644 to 0.682, with a relatively small gain due to the maturity of modern 2D detectors. For 3D detection, overall AP rises from 0.222 to 0.295. For trajectory prediction, ADE decreases from 1.339 to 0.812, whereas direct regression reaches 5.391. For semantic segmentation, the fusion pipeline raises mIoU from 0.599 to 0.742. Complete AP/AR, category-wise AP, ADE/FDE, precision/recall, and absolute GFLOPs are provided in Appendix~\ref{app:full-results}.

\begin{table*}[!t]
\centering
\caption{Representative evidence across four structured-prediction tasks. Relative cost is reported as specialist/direct regression/selection. The comparison is between collaboration paradigms: the proxy learner receives specialist candidates, whereas the direct regressor does not. Complete task-specific metrics and absolute GFLOPs are reported in Appendix~\ref{app:full-results}.}
\label{tab:main-results}
\small
\setlength{\tabcolsep}{6.0pt}
\renewcommand{\arraystretch}{1.08}
\begin{tabularx}{\textwidth}{@{}Y l c c c c@{}}
\toprule
Task & Metric & Specialist & Direct VLM & Selector & Relative cost \\
\midrule
2D object detection & AP@0.50 $\uparrow$ & 0.644 & 0.554 & \best{0.682} & $1.0/4.9/1.4\times$ \\
3D object detection & AP $\uparrow$ & 0.222 & 0.148 & \best{0.295} & $1.0/5.1/1.5\times$ \\
Trajectory prediction & ADE $\downarrow$ & 1.339 & 5.391 & \best{0.812} & $1.0/4.3/1.5\times$ \\
Semantic segmentation & mIoU $\uparrow$ & 0.599 & 0.544 & \best{0.742} & $1.0/7.1/1.8\times$ \\
\bottomrule
\end{tabularx}
\end{table*}

\subsection{Ablation studies}

\subsubsection{Source of gains}

Table~\ref{tab:source-ablation} separates candidate coverage from selection quality. Random selection uses the same candidate sets and is averaged over five seeds. Confidence selection uses specialist confidences or mode probabilities when available; perturbed candidates inherit the parent confidence with a magnitude penalty. Heuristic reranking uses projected alignment and aspect-ratio priors for boxes, curvature and lane-compliance penalties for trajectories, and overlap/coverage rules for mask partitions. The oracle selects the ground-truth-best candidate and therefore serves only as an upper bound on candidate-set coverage, rather than a deployable method.

The results show that proxy reasoning over candidate structures brings clear improvements, but also that choosing the best candidate remains nontrivial. Random selection generally fails to match the specialist, indicating that candidate expansion alone does not explain the gains. Confidence-based and heuristic selection recover part of the available improvement, while LoRA adaptation and explicit comparative reasoning further enhance selection quality. Nevertheless, the remaining gap to the oracle indicates that even a learned proxy reasoner cannot always identify the best candidate from the set, leaving room for stronger comparative reasoning and candidate-ranking mechanisms.

\begin{table*}[!t]
\centering
\caption{Source-of-gain ablation. Bold marks the best deployable variant; the oracle is a non-deployable upper bound over the same candidate set.}
\label{tab:source-ablation}
\footnotesize
\setlength{\tabcolsep}{6.2pt}
\renewcommand{\arraystretch}{1.06}
\begin{tabularx}{\textwidth}{@{}Y c c c c@{}}
\toprule
Variant & 2D AP$_{50}$ $\uparrow$ & 3D AP $\uparrow$ & Traj. ADE $\downarrow$ & Seg. mIoU $\uparrow$ \\
\midrule
Specialist only & 0.644 & 0.222 & 1.339 & 0.599 \\
Random over candidates & 0.612 & 0.203 & 1.520 & 0.610 \\
Confidence selection & 0.648 & 0.228 & 1.190 & 0.651 \\
Heuristic reranking & 0.651 & 0.235 & 0.980 & 0.704 \\
Base VLM selector, no LoRA & 0.649 & 0.242 & 0.974 & 0.712 \\
LoRA selector, no comparative reasoning & 0.663 & 0.253 & 0.878 & 0.725 \\
\method, LoRA + comparative reasoning & \best{0.682} & \best{0.295} & \best{0.812} & \best{0.742} \\
Oracle best-of-$K$ & 0.711 & 0.322 & 0.594 & 0.803 \\
\bottomrule
\end{tabularx}
\end{table*}

\subsubsection{Design sensitivity}

Candidate-set size and perturbation scale control the balance between hypothesis coverage and proxy-task difficulty. Table~\ref{tab:design-ablation} shows rapid improvement from $K=1$ to $K=10$, followed by saturation at the largest task-dependent set ($K=15$ or $20$). We therefore use $K=10$ as the main operating point: larger sets provide only marginal gains while increasing prompt length and visual clutter. The base perturbation scale, 1.000$\times$, performs best across all four tasks; smaller perturbations provide insufficient diversity, whereas larger perturbations introduce implausible distractors.

The same table also analyzes supervision and information-space reconstruction. The selector already recovers most of its final performance with only a limited fraction of the fine-tuning data, and performance largely saturates as the data scale increases. This suggests that the proxy task is data-efficient because the foundation model only needs to learn comparative reasoning over structured alternatives, rather than the full metric prediction task. Prompt representation further confirms the importance of reconstruction: text-only proxy reasoning provides limited gains, whereas image-based overlays make the geometric consequences of specialist hypotheses directly visible to the foundation model. These results indicate that the effectiveness of \framework depends not only on candidate generation, but also on reconstructing candidates into a representation where contextual comparison becomes easy.

\begin{table*}[!t]
\centering
\caption{Unified design ablations. $\dagger$ marks the main $K=10$ operating point; bold denotes the best value within each ablation block.}
\label{tab:design-ablation}
\footnotesize
\setlength{\tabcolsep}{6.0pt}
\renewcommand{\arraystretch}{1.03}
\begin{tabularx}{\textwidth}{@{}Y c c c c@{}}
\toprule
Setting & 2D AP$_{50}$ $\uparrow$ & 3D AP $\uparrow$ & Traj. ADE $\downarrow$ & Seg. mIoU $\uparrow$ \\
\midrule
\multicolumn{5}{@{}l}{\textit{Candidate-set size $K$}} \\
$K=1$ & 0.644 & 0.222 & 1.339 & 0.599 \\
$K=3$ & 0.654 & 0.242 & 1.012 & 0.702 \\
$K=6$ & 0.673 & 0.274 & 0.871 & 0.731 \\
$K=10^{\dagger}$ & \best{0.682} & 0.295 & 0.812 & 0.742 \\
$K=15/20$ & \best{0.682} & \best{0.295} & \best{0.805} & \best{0.745} \\
\addlinespace[0.22em]
\multicolumn{5}{@{}l}{\textit{Perturbation scale}} \\
Low, 0.500$\times$ & 0.657 & 0.264 & 0.925 & 0.721 \\
Base, 1.000$\times$ & \best{0.682} & \best{0.295} & \best{0.812} & \best{0.742} \\
High, 1.500$\times$ & 0.654 & 0.276 & 0.865 & 0.736 \\
\addlinespace[0.22em]
\multicolumn{5}{@{}l}{\textit{Selector fine-tuning data}} \\
10\% & 0.678 & 0.291 & 0.819 & 0.739 \\
25\% & 0.681 & 0.293& 0.813 & 0.742 \\
50\% & 0.682 & 0.294 & 0.813 & 0.741 \\
100\% & \best{0.682} & \best{0.295} & \best{0.812} & \best{0.742} \\
\addlinespace[0.22em]
\multicolumn{5}{@{}l}{\textit{Prompt representation}} \\
Text + comparative reasoning & 0.646 & 0.223 & 1.332 & 0.604 \\
Image Overlay + comparative reasoning & \best{0.682} & \best{0.295} & \best{0.812} & \best{0.742} \\
\bottomrule
\end{tabularx}
\end{table*}

\subsection{Qualitative analysis}

\begin{figure}[H]
  \centering
  \includegraphics[width=\columnwidth]{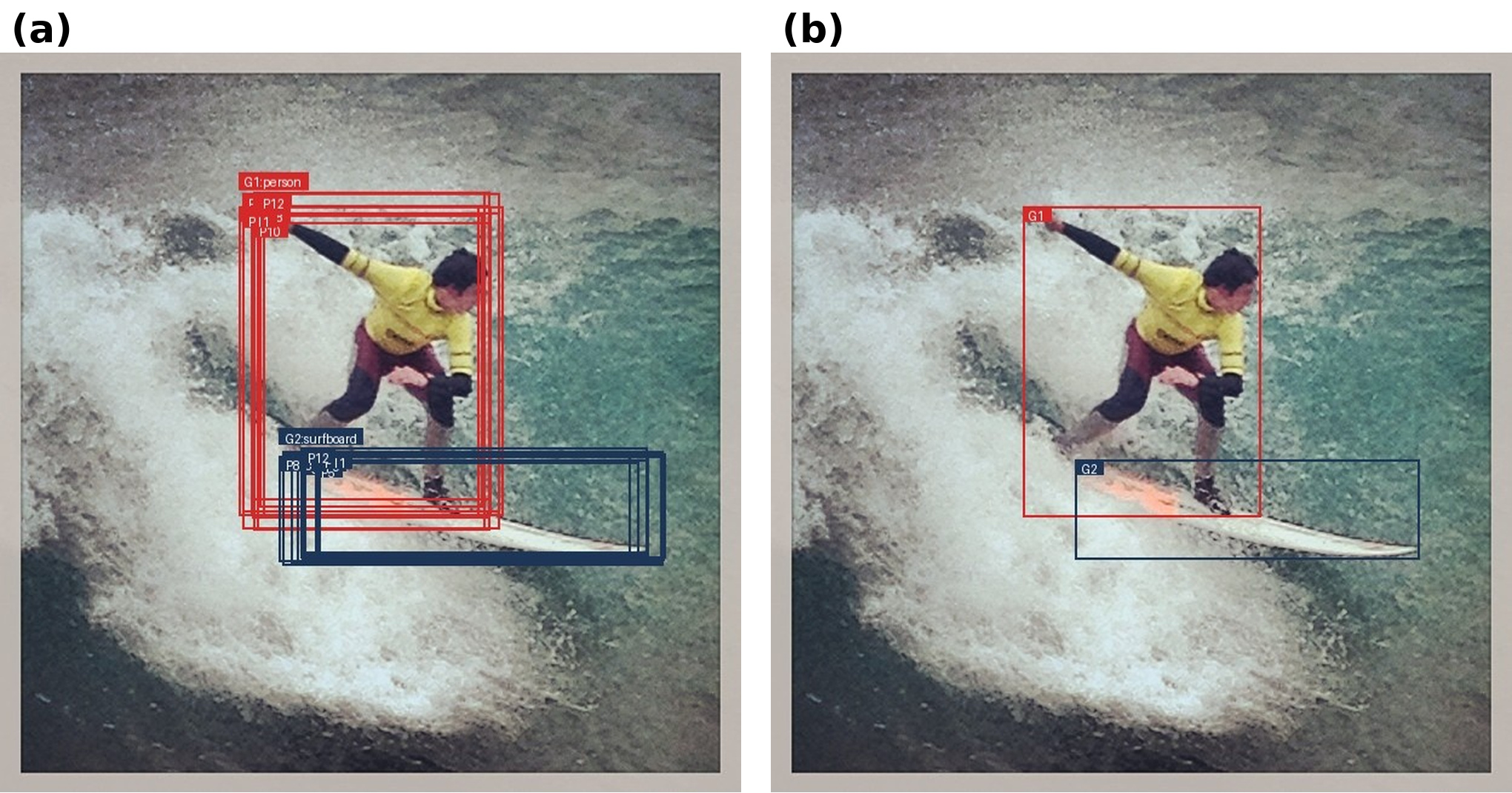}
  \captionsetup{font=footnotesize,skip=2pt}
  \caption{2D detection in a surfer scene. Specialist proposals and local perturbations form the candidate set; the VLM selects one existing box for each target group.}
  \label{fig:qual-2d}
\end{figure}

\begin{figure}[H]
  \centering
  \includegraphics[width=\columnwidth]{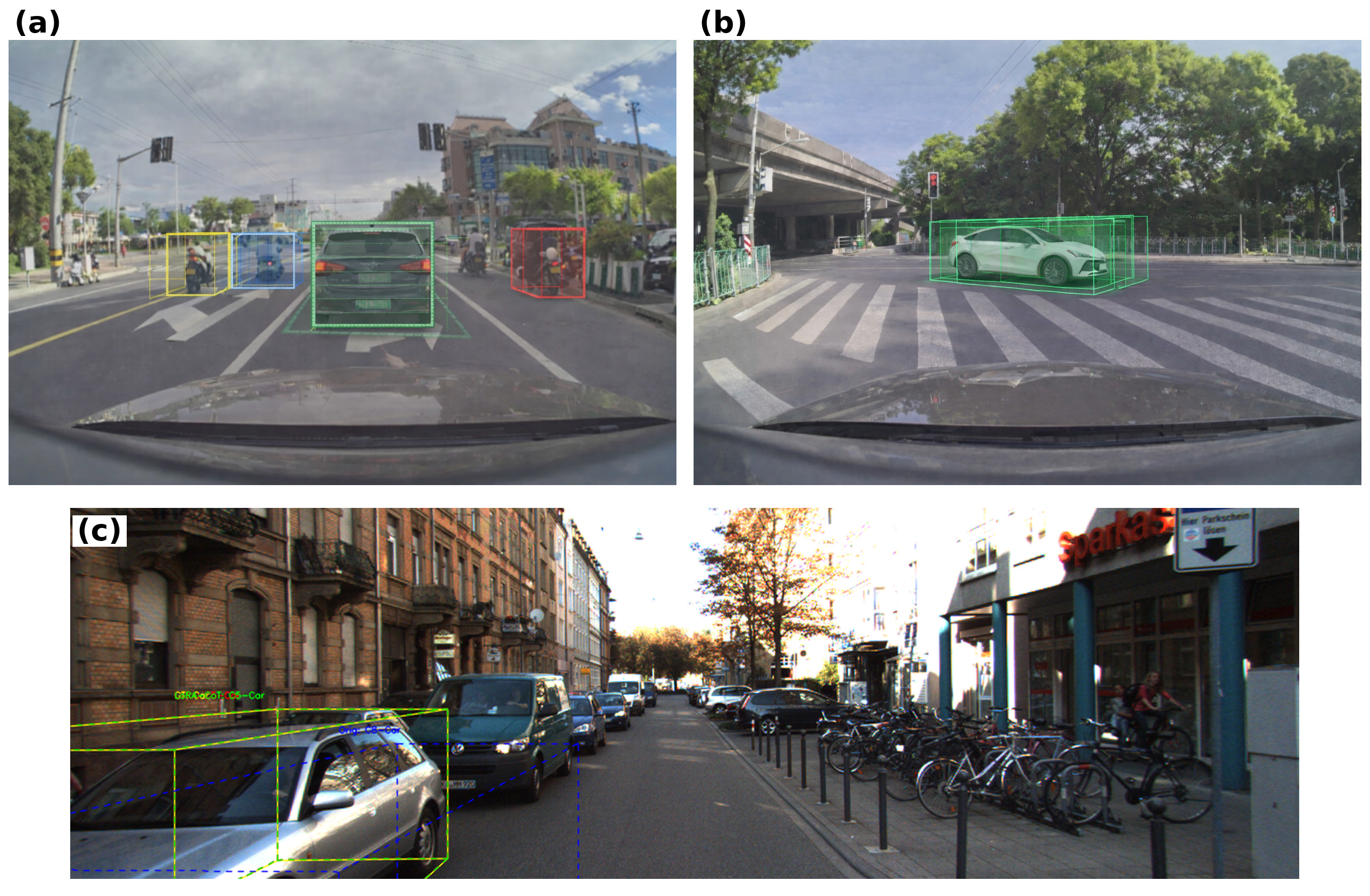}
  \captionsetup{font=footnotesize,skip=2pt}
  \caption{Qualitative 3D detection. Panels (a) and (b) show selected projected boxes; in panel (c), green denotes \method and blue denotes direct VLM regression.}
  \label{fig:qual-3d}
\end{figure}

\begin{figure}[H]
  \centering
  \includegraphics[width=\columnwidth]{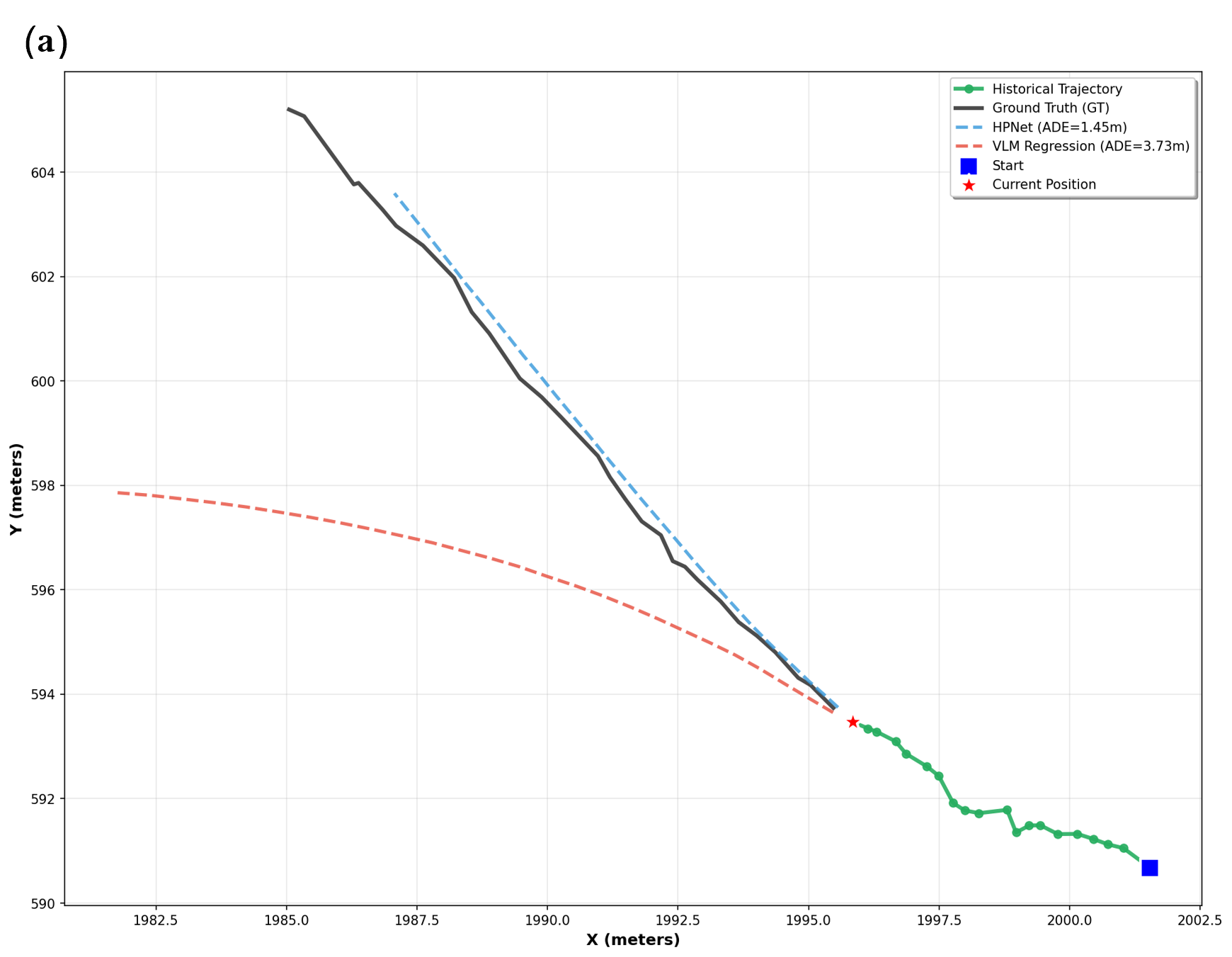}

  \vspace{0.3em}

  \includegraphics[width=\columnwidth]{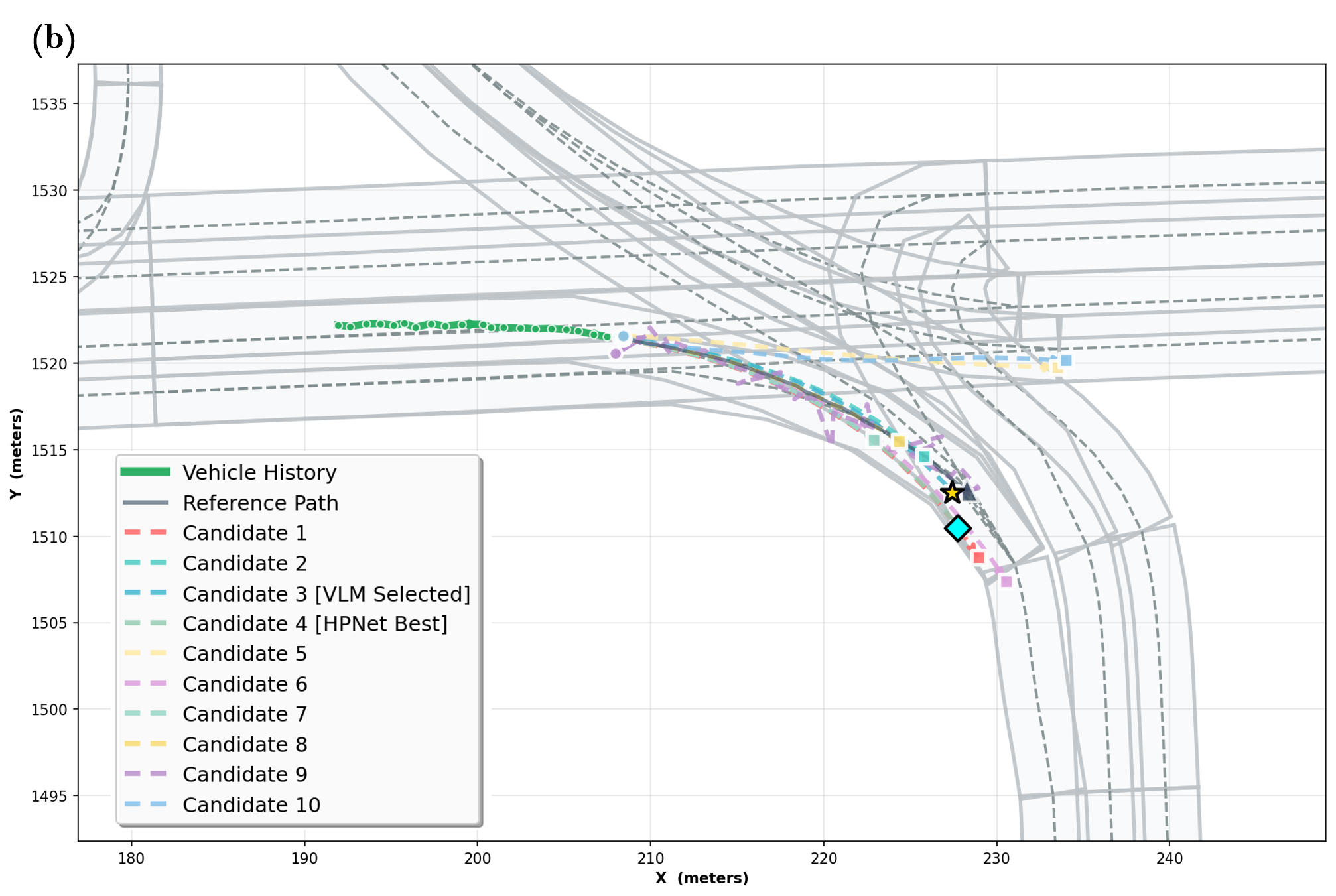}

  \captionsetup{font=footnotesize,skip=2pt}
  \caption{Trajectory prediction. Panel (a) compares direct VLM regression with HPNet against the annotated future, illustrating endpoint deviation and continuity errors in direct regression. Panel (b) shows proxy selection over HPNet modes and perturbed candidates in map context. The gray reference path is used only as an inference-time reference and is distinct from the annotated future used for supervision and evaluation.}
  \label{fig:qual-traj}
\end{figure}

\begin{figure}[H]
  \centering
  \includegraphics[width=\columnwidth]{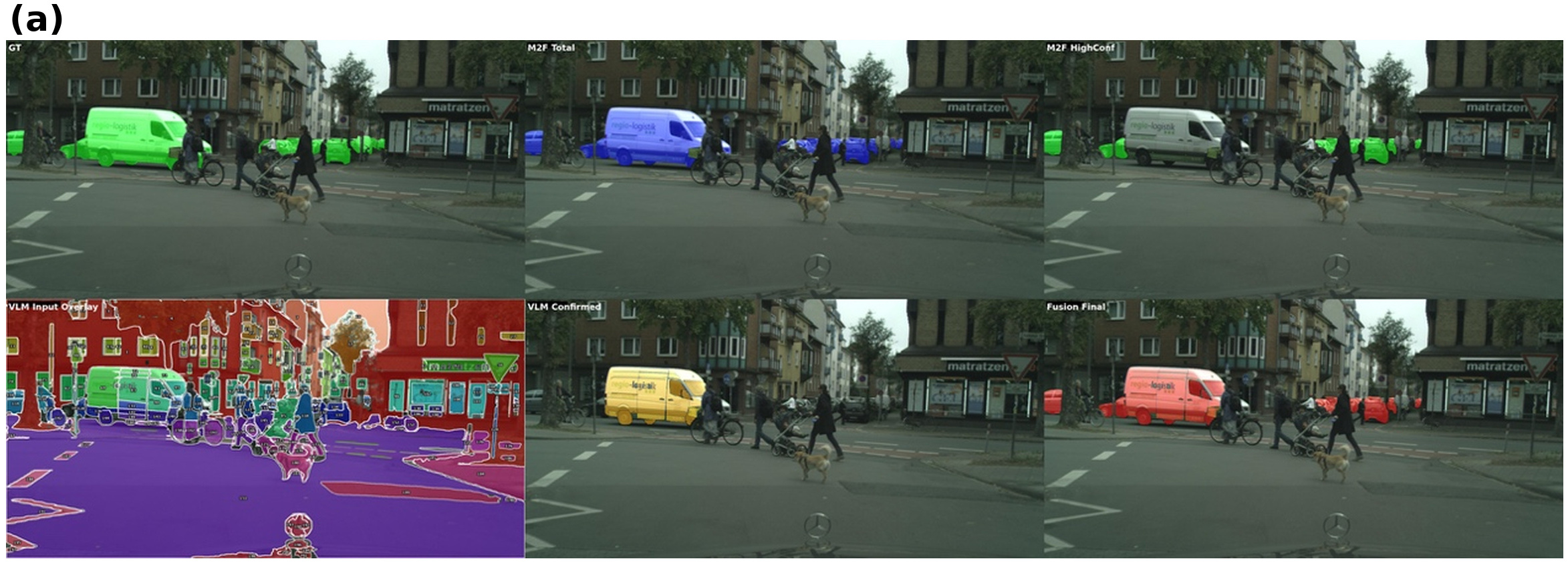}

  \vspace{0.3em}

  \includegraphics[width=\columnwidth]{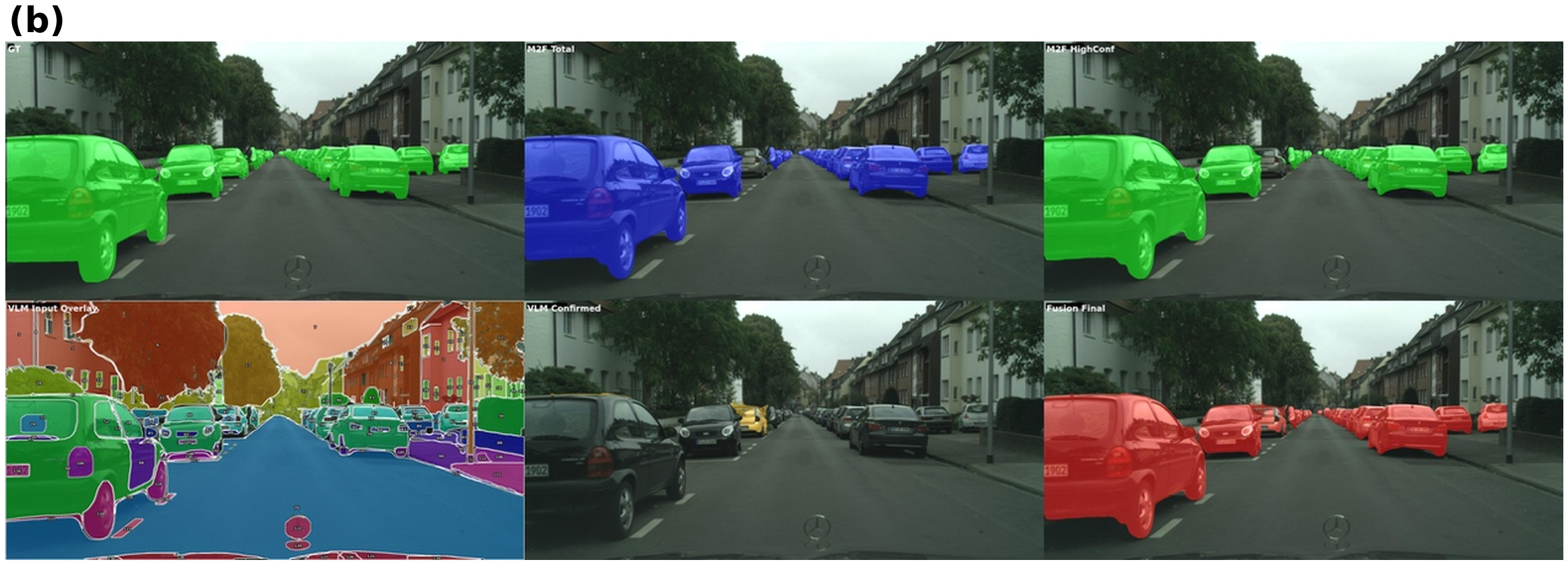}

  \captionsetup{font=footnotesize,skip=2pt}
  \caption{Cityscapes semantic segmentation. Panels (a) and (b) show two representative examples. Each example includes the ground-truth car mask, Mask2Former predictions, numbered FastSAM partitions, VLM-confirmed regions, and the fused result.}
  \label{fig:qual-seg}
\end{figure}

The task-level visualizations in Figs.~\ref{fig:qual-2d}--\ref{fig:qual-seg} illustrate how the proxy reasoner complements rather than replaces the specialist. These examples are qualitative evidence rather than substitutes for quantitative evaluation, but they provide intuitive evidence of the failure modes addressed by bounded comparative reasoning. Across all tasks, the foundation model does not freely synthesize new coordinates, masks, or trajectories. Instead, it evaluates the visual and contextual consequences of specialist-generated hypotheses and returns a bounded decision, such as a candidate index or a set of verified region identifiers.

\paragraph{Detection.}
Fig.~\ref{fig:qual-2d} shows a representative 2D detection case in which detector proposals and local perturbations produce several plausible boxes with different extents, alignments, and overlaps with the visible targets. The selected result suppresses redundant or poorly aligned alternatives while retaining detector-generated coordinates for both the person and the surfboard. Fig.~\ref{fig:qual-3d} further illustrates the role of image-conditioned comparison in 3D detection. Even when candidates share the same semantic class, their projected boxes may differ in apparent depth, yaw, object extent, and alignment with image evidence. Because the proxy decision returns an existing candidate identifier, the final output preserves a valid box parameterization rather than relying on free-form numerical decoding by the VLM.

\paragraph{Trajectory prediction.}
The trajectory examples in Fig.~\ref{fig:qual-traj} illustrate two complementary aspects of the proposed proxy-task formulation. Fig.~\ref{fig:qual-traj}(a) compares direct VLM regression with HPNet and shows that direct regression can suffer from endpoint deviation and reduced continuity with the observed motion, especially toward the prediction horizon. In contrast, Fig.~\ref{fig:qual-traj}(b) visualizes proxy selection over HPNet modes and perturbed candidates in map context. By rendering candidate futures together with the observed history and lane geometry, the proxy task makes continuity, lane compliance, and destination plausibility directly comparable. The returned trajectory is therefore selected from a precomputed, structurally valid candidate set. The gray reference path is an inference-time reference and is distinct from the annotated future used for supervision and evaluation.

\paragraph{Semantic segmentation.}
Fig.~\ref{fig:qual-seg} demonstrates the division of labor between semantic priors and boundary support in Cityscapes. High-confidence Mask2Former regions provide category-aware semantic predictions, while FastSAM supplies atomic partitions that preserve local boundary structure. The VLM then verifies which numbered regions are associated with the target car category. This design allows the final fusion to recover disconnected or partially missed target regions without requiring the VLM to generate pixels, polygons, or contours directly. Across the four visualized tasks, the same pattern emerges: task-specific specialists preserve geometric, dynamic, and dense spatial validity, while the foundation model contributes contextual comparison and proxy-level reasoning over bounded alternatives.

\section{Discussion and Limitations}

\paragraph{Why proxy-task reasoning helps.}
Direct regression maps an observation into a large continuous output space and requires precise numerical calibration, valid serialization, and task-specific structural consistency. Proxy-task reasoning converts this open-ended prediction problem into bounded reasoning over viable hypotheses that have already been generated by a specialist. The foundation model can therefore focus on semantic consistency, contextual comparison, and ambiguity resolution without relearning the full geometry, dynamics, or dense spatial structure of the original task. The oracle gaps in Table~\ref{tab:source-ablation} show that candidate coverage is necessary, whereas the differences among random selection, heuristic reranking, and learned proxy reasoning show that coverage alone is not sufficient.

\paragraph{Not a conventional ensemble.}
Standard ensembles often combine predictions through fixed averaging, voting, or confidence-based rules, and their constituent models need not inspect one another's outputs. In \method, the foundation model explicitly conditions on the original scene and on the reconstructed visual consequences of each specialist-generated hypothesis. It evaluates what each candidate implies in context and returns a bounded proxy decision, such as a candidate identifier or a set of verified regions. This preserves the structured output space of the specialist instead of averaging incompatible predictions or asking the foundation model to synthesize new geometry.

\paragraph{Limitations.}
The proposed framework cannot recover a correct answer if it is absent from the candidate set. Crowded overlays may cause index confusion, and semantically salient objects may attract attention even when their geometric alignment is inferior. Candidate generators and foundation models may also share dataset or scene biases, so proxy reasoning does not guarantee independent error correction. Our experiments use one VLM backbone and a limited set of specialist models, and they evaluate open-loop task metrics rather than closed-loop driving or robotic safety. Future work should report calibrated selection confidence, invalid-output rates, oracle-conditioned accuracy, robustness to adversarial near-duplicate candidates, distribution-shift performance, and closed-loop effects. A deterministic specialist fallback remains necessary for deployment, but it is not a substitute for task-specific validation, confidence calibration, and safety testing. The same bounded-authority interface is also compatible with human-in-the-loop correction and takeover mechanisms \citep{wu2023humanloop}.

\section{Conclusion}

We presented \framework, a role-aligned foundation-model collaboration framework that enhances specialized models through proxy-task reasoning, and instantiated it as \method for structured prediction. Across four tasks, this allocation improves specialist accuracy, substantially outperforms direct foundation-model regression, and preserves task-specific geometric and physical constraints. The ablations show that strong performance requires both useful candidate coverage and learned multimodal comparison. More broadly, our results support a practical design rule for embodied intelligence systems: specialized models should construct valid hypotheses, while foundation models should reason over bounded proxy tasks to refine them.

% Balance the final two-column bibliography page before switching to the appendix.
\balance
\printbibliography[title={References}]
\end{refsection}

\clearpage
\onecolumn
\appendix
\setcounter{figure}{0}
\renewcommand{\thefigure}{A\arabic{figure}}
\setcounter{table}{0}
\renewcommand{\thetable}{A\arabic{table}}
\setcounter{equation}{0}
\renewcommand{\theequation}{A\arabic{equation}}
\raggedbottom
\setlength{\parskip}{0.10em}
\setlength{\textfloatsep}{6pt plus 1pt minus 1pt}
\setlength{\floatsep}{6pt plus 1pt minus 1pt}
\setlength{\intextsep}{6pt plus 1pt minus 1pt}
\titlespacing*{\section}{0pt}{0.85ex plus .15ex minus .1ex}{0.35ex}
\titlespacing*{\subsection}{0pt}{0.65ex plus .12ex minus .1ex}{0.25ex}

\begin{center}
{\LARGE\bfseries Appendix}
\end{center}
\vspace{0.25em}
\noindent This appendix reports the complete task-specific metrics and absolute inference costs omitted from the compact main-text comparison, followed by implementation details and the full selector and direct-regression prompts.

\section{Complete Experimental Results}\label{app:full-results}

Table~\ref{tab:main-results} retains one primary metric and normalized cost per task for readability. Tables~\ref{tab:app-2d}--\ref{tab:app-seg} report every measured task-specific metric and the corresponding absolute inference cost. Performance values are shown to three decimals and GFLOPs per sample are rounded to integers.

\subsection{Two-dimensional object detection}

\begin{table}[H]
\centering
\caption{Two-dimensional detection performance on COCO 2017.}
\label{tab:app-2d}
\small
\setlength{\tabcolsep}{5.0pt}
\renewcommand{\arraystretch}{1.10}
\begin{tabularx}{\linewidth}{@{}Y c c c c c@{}}
\toprule
Method & AP (0.50:0.95) & AR (0.50:0.95) & AP@0.50 & AR@0.50 & GFLOPs/sample \\
\midrule
RT-DETR & 0.485 & 0.374 & 0.644 & 0.580 & 198 \\
Direct VLM regression & 0.307 & 0.230 & 0.554 & 0.507 & 966 \\
\method & \best{0.542} & \best{0.440} & \best{0.682} & \best{0.618} & 270 \\
\bottomrule
\end{tabularx}
\end{table}

ProxySelect improves all reported AP and AR values over RT-DETR. The gain is smaller than in the other tasks, consistent with the maturity and strong inductive bias of modern 2D detectors.

\subsection{Three-dimensional object detection}

\begin{table}[H]
\centering
\caption{Category-wise three-dimensional detection performance on KITTI.}
\label{tab:app-3d}
\small
\setlength{\tabcolsep}{4.6pt}
\renewcommand{\arraystretch}{1.10}
\begin{tabularx}{\linewidth}{@{}Y c c c c c@{}}
\toprule
Method & Pedestrian AP@50 & Car AP@50 & Cyclist AP@50 & Overall AP & GFLOPs/sample \\
\midrule
3D-MOOD & 0.261 & 0.623 & 0.222 & 0.222 & 405 \\
Direct VLM regression & 0.174 & 0.501 & 0.167 & 0.148 & 2050 \\
\method & \best{0.323} & \best{0.685} & \best{0.297} & \best{0.295} & 620 \\
\bottomrule
\end{tabularx}
\end{table}

Selection improves all three reported categories and raises overall AP from 0.222 to 0.295. AP@50 denotes class-specific average precision at an intersection-over-union threshold of 0.50; overall AP follows the reported 3D-MOOD evaluation protocol.

\subsection{Trajectory prediction}

\begin{table}[H]
\centering
\caption{Trajectory prediction performance on Argoverse.}
\label{tab:app-traj}
\small
\setlength{\tabcolsep}{8.0pt}
\renewcommand{\arraystretch}{1.10}
\begin{tabularx}{\linewidth}{@{}Y c c c@{}}
\toprule
Method & ADE $\downarrow$ & FDE $\downarrow$ & GFLOPs/sample \\
\midrule
HPNet & 1.339 & 2.850 & 270 \\
Direct VLM regression & 5.391 & 10.309 & 1170 \\
\method & \best{0.812} & \best{1.660} & 418 \\
\bottomrule
\end{tabularx}
\end{table}

The evaluation set contains 10,000 trajectories. Selection over HPNet modes and perturbations reduces both average and final displacement error, while direct regression exhibits substantially larger endpoint error.

\subsection{Semantic segmentation}

\begin{table}[H]
\centering
\caption{Semantic segmentation performance on the Cityscapes validation set for the car category.}
\label{tab:app-seg}
\small
\setlength{\tabcolsep}{7.0pt}
\renewcommand{\arraystretch}{1.10}
\begin{tabularx}{\linewidth}{@{}Y c c c c@{}}
\toprule
Method & mIoU $\uparrow$ & mPrecision $\uparrow$ & mRecall $\uparrow$ & GFLOPs/sample \\
\midrule
Mask2Former & 0.599 & 0.632 & \best{0.929} & 265 \\
Direct VLM regression & 0.544 & \best{0.819} & 0.592 & 1877 \\
Mask2Former + FastSAM + VLM & \best{0.742} & 0.792 & 0.918 & 489 \\
\bottomrule
\end{tabularx}
\end{table}

Mask2Former provides high recall but incomplete high-confidence coverage. Direct mask regression is more precise but loses substantial recall. The selection pipeline preserves specialist geometry while recovering semantically valid atomic regions, yielding the strongest mIoU.

\section{Implementation Details}

\subsection{Candidate labels, randomization, and parsing}
For 2D and 3D detection, the target candidate is defined by the highest task-matched overlap with the annotation. For trajectory prediction, the target minimizes displacement error. Semantic segmentation uses the set of atomic regions whose union best matches the target category. Candidate identifiers are randomly permuted before every rendered sample, and labels are remapped after permutation; no candidate source therefore occupies a fixed index.

The direct baseline decodes deterministic, fixed output schemas. Invalid JSON, out-of-range coordinates, malformed polygons, or an incorrect number of trajectory waypoints are treated as parse failures. The selector parser extracts the final candidate identifier, one identifier per 2D target group, or a comma-separated list of segmentation-region identifiers. In deployment, invalid selector output should trigger the unmodified specialist prediction.

\subsection{Task-specific candidate construction}
\paragraph{2D object detection.}
The original selection template lists six candidates. When additional YOLO-family proposals are available, their records are appended in the same format. The prompt's use of ``IoU'' refers to qualitative visual overlap with the visible target; no ground-truth IoU is available at inference.

\paragraph{3D object detection.}
Candidate descriptions include class, 3D location, dimensions, and yaw. Training includes category perturbations to prevent the selector from relying only on geometry. At inference, all candidates derive from the 3D-MOOD output.

\paragraph{Trajectory prediction.}
Candidate descriptions summarize continuity, endpoint plausibility relative to the gray inference-time reference, driving realism, and road compliance. The annotated future is used only to construct training labels and compute ADE/FDE. It is never shown to the selector.

\paragraph{Semantic segmentation.}
Mask2Former high-confidence regions pass through unchanged. FastSAM atomic partitions that are largely covered by the high-confidence prior or contain too little uncovered area are filtered before VLM verification. The VLM returns region indices rather than pixels or polygon vertices.

\section*{Datasets and Responsible Use}
The experiments use KITTI, Argoverse, COCO 2017, and Cityscapes. These benchmarks are used under their respective licenses. The proposed proxy-task framework is not a safety-certified component. Its outputs should be validated through task-specific constraints, confidence calibration, fallback behavior, and closed-loop testing before any real-world deployment.

\section{Prompt and Output Specifications}

The following templates reproduce the selector and direct-regression interfaces. Candidate fields are populated with task-specific values at runtime. Candidates are randomly permuted and renumbered before each query, and target labels are remapped after permutation.

\subsection{Selection prompts}

\begin{promptbox}{2D object detection -- selection prompt}
Analyze the following 2D bounding box candidates and select the most accurate one step-by-step:
Candidate 1: Class: xxx, Position: (x=..., y=...), Size: (width:..., height=...), Area:...
Candidate 2: Class: xxx, Position: (x=..., y=...), Size: (width:..., height=...), Area:...
...
Candidate 6: Class: xxx, Position: (x=..., y=...), Size: (width:..., height=...), Area:...
Please follow these steps:
1. Analyze the position of each candidate (horizontal and vertical alignment)
2. Analyze the size and aspect ratio of each candidate
3. Evaluate the overlap (IoU) with the target object
4. Verify the class accuracy of each candidate
5. Select the most suitable candidate based on comprehensive analysis
Finally answer: Candidate X is the best choice.
\end{promptbox}

\begin{promptbox}{3D object detection -- selection prompt}
Analyze the following 3D bounding box candidates and select the most accurate one through step-by-step analysis of each candidate's characteristics:
Candidate 0: class xx, location x y z, dimensions w h l, rotation_y yaw
Candidate 1: class xx, location x y z, dimensions w h l, rotation_y yaw
...
Candidate 9: class xx, location x y z, dimensions w h l, rotation_y yaw
Please follow these steps for analysis:
1. Analyze the position characteristics of each candidate
2. Analyze the size characteristics of each candidate
3. Analyze the orientation characteristics of each candidate
4. Based on the above analysis, select the most suitable candidate
Finally answer: Candidate X is the best choice.
\end{promptbox}

\begin{promptbox}{Trajectory prediction -- selection prompt}
You are analyzing a trajectory prediction scenario. The image shows:
- Green solid line with dots: vehicle historical trajectory
- Gray lines: road lane boundaries and centerlines
- Gray line: reference trajectory for comparison
- Colored dashed lines: future trajectory candidates
Your task is to select the BEST trajectory among the candidates.
Evaluation Criteria:
1. Visual Continuity: Does it smoothly continue from the historical path?
2. Endpoint Accuracy: Does it reach a reasonable destination based on the reference?
3. Driving Realism: Would a human driver naturally follow this path?
4. Road Compliance: Does it follow the road structure shown in gray?
Select the candidate that best balances these criteria.
Respond with ONLY the number (e.g., "Candidate 3" or "3").
\end{promptbox}

\begin{promptbox}{Semantic segmentation -- selection prompt}
Query category: <TARGET_CATEGORY>
The image shows N candidate masks outlined in different colors and numbered 1-N. Select all masks that belong to the query category. If there are multiple, separate them with commas; if none, answer None.
\end{promptbox}

\subsection{Direct-regression prompts and output schemas}

For the system-level comparison, the direct baseline receives the raw task input and a deterministic output specification but no specialist predictions or candidate overlays.

\begin{promptbox}{2D object detection -- direct-regression prompt}
Detect all requested objects in the image and directly predict their image-coordinate bounding boxes.
Return valid JSON only:
{"objects":[{"class":"<class>","x1":<float>,"y1":<float>,"x2":<float>,"y2":<float>}]}
Requirements:
- coordinates must satisfy x1 < x2 and y1 < y2 and remain inside the image bounds;
- use the requested class vocabulary;
- do not add markdown or explanatory text;
- if no requested object is visible, return {"objects":[]}.
\end{promptbox}

\begin{promptbox}{3D object detection -- direct-regression prompt}
Detect the requested objects in the camera image and directly predict one 3D bounding box per object using the dataset coordinate convention.
Return valid JSON only:
{"objects":[{"class":"<class>","x":<float>,"y":<float>,"z":<float>,"w":<float>,"h":<float>,"l":<float>,"yaw":<float>}]}
Requirements:
- use the requested class vocabulary;
- use finite numerical values only;
- do not add markdown or explanatory text;
- if no requested object is visible, return {"objects":[]}.
\end{promptbox}

\begin{promptbox}{Trajectory prediction -- direct-regression prompt}
Given the agent's observed history and the map or lane context, directly predict T future positions in the local coordinate system at the specified time interval.
Return valid JSON only:
{"trajectory":[[x1,y1],[x2,y2],...,[xT,yT]]}
Requirements:
- output exactly T ordered waypoints;
- the first point must continue from the observed history;
- use finite numerical values only;
- do not output multiple modes, prose, markdown, or confidence values.
\end{promptbox}

\begin{promptbox}{Semantic segmentation -- direct-regression prompt}
Segment every visible instance of <TARGET_CATEGORY> directly from the original image. Return valid JSON only:
{"category":"<TARGET_CATEGORY>","polygons":[[[x1,y1],[x2,y2],...,[xn,yn]],...]}
Requirements: each polygon has at least three vertices; vertices follow the visible boundary and do not self-intersect; use multiple polygons for disconnected components; exclude background and other categories; return {"category":"<TARGET_CATEGORY>","polygons":[]} if absent; do not add prose, markdown, or confidence values. For evaluation, polygons are rasterized at the original resolution; invalid JSON, out-of-range coordinates, or invalid polygons are parse failures.
\end{promptbox}

\enlargethispage{2.1cm}
\end{document}